\documentclass[lettersize,journal,twoside]{IEEEtran}

\usepackage{orcidlink}
\usepackage{algorithm}
\usepackage{algorithmic}
\usepackage{cite}  

\usepackage{multirow}
\usepackage[utf8]{inputenc} % allow utf-8 input
\usepackage[T1]{fontenc}    % use 8-bit T1 fonts
\usepackage{url}            % simple URL typesetting
\usepackage{booktabs}       % professional-quality tables
\usepackage{amsfonts}       % blackboard math symbols
\usepackage{nicefrac}       % compact symbols for 1/2, etc.
\usepackage{microtype}      % microtypography
\usepackage[table]{xcolor} 
\usepackage{amsmath}
\usepackage{adjustbox}
\usepackage{subcaption}
\usepackage{enumitem}
\definecolor{ImportantColor}{RGB}{211, 222, 190}

\newcommand{\SE}{\mathrm{SE}}
\newcommand{\SO}{\mathrm{SO}}

\newcommand{\ac}{\mathbf{A}}
\newcommand{\ob}{\mathbf{O}}
\newcommand{\as}{\mathbf{a}}
\newcommand{\tb}[3]{\setlength{\tabcolsep}{#2mm}\begin{tabular}{#1}#3\end{tabular}}

\begin{document}
%Adaptive Diffusion Policies via Geometry-Aware Denoising and Task-Aware Initialization
%Manifold-Constrained Diffusion for Test-Time Adaptation in Robotic Manipulation
\title{ADPro: a Test-time Adaptive Diffusion Policy via Manifold-constrained Denoising and Task-aware Initialization for Robotic Manipulation}%
%\title{ADPro: a Test-time Adaptive Diffusion Policy with Unified Manifold Constraints for Robotic Manipulation}%

% \author{\IEEEauthorblockN{Anonymous Authors}}

\author{Zezeng Li\orcidlink{0000-0001-9064-689X},
        Rui Yang\orcidlink{0000-0002-2102-4306},
        Ruochen Chen\orcidlink{0009-0001-0036-925X}, 
        ZhongXuan Luo\orcidlink{0000-0001-5997-2646}, and~Liming Chen\orcidlink{0000-0002-3654-9498},~\IEEEmembership{Senior~Member,~IEEE,}% <-this % stops a space
\thanks{Zezeng Li, Rui Yang, Ruochen Chen, and Liming Chen are with École Centrale de Lyon, France. \\ ZhongXuan Luo is with Dalian University of Technology, China.}%
}


\markboth{IEEE Transactions on Neural Networks and Learning Systems. Preprint Version}%
{Li \MakeLowercase{\textit{et al.}}: a Test-time Adaptive Diffusion Policy via Manifold-constrained Denoising and Task-aware Initialization for Robotic}



\maketitle

\begin{abstract}
Diffusion policies have recently emerged as a powerful class of visuomotor controllers for robot manipulation, offering stable training and expressive multi-modal action modeling. However, existing approaches typically treat action generation as an unconstrained denoising process, ignoring valuable a priori knowledge about geometry and control structure. In this work, we propose the Adaptive Diffusion Policy (ADP), a test-time adaptation method that introduces two key inductive biases into the diffusion. First, we embed a geometric manifold constraint that aligns denoising updates with task-relevant subspaces, leveraging the fact that the relative pose between the end-effector and target scene provides a natural gradient direction, and guiding denoising along the geodesic path of the manipulation manifold. Then, to reduce unnecessary exploration and accelerate convergence, we propose an analytically guided initialization: rather than sampling from an uninformative prior, we compute a rough registration between the gripper and target scenes to propose a structured initial noisy action. ADP is compatible with pre-trained diffusion policies and requires no retraining, enabling test-time adaptation that tailors the policy to specific tasks, thereby enhancing generalization across novel tasks and environments. Experiments on RLBench, CALVIN, and real-world dataset show that ADPro, an implementation of ADP, improves success rates, generalization, and sampling efficiency, achieving up to 25\% faster execution and 9\% points over strong diffusion baselines.
\end{abstract}


\section{Introduction}
Autonomous robotic manipulation in unstructured environments requires policies that not only predict accurate actions but also generalize well across diverse tasks and scenes. Early approaches based on rule-based controllers ~\cite{miller2004graspit,li2007data,dang2011blind,sun2021adaptive} or dense discriminative models ~\cite{ten2017grasp,liang2019pointnetgpd,mousavian20196,murali20206,fang2020graspnet,bing2021robotic,gou2021rgb,zhao2021regnet,depierre2021scoring,wei2021gpr,ma2023towards,wei2023discriminative,qin2023rgb,ryuequivariant,vahabpour2024diverse} have struggled to scale, either due to lack of flexibility or the computational cost of exhaustive search. More recently, \textbf{diffusion policies (DPs)} ~\cite{chi2023diffusion,wei2025grasp,he2025learning,wangadamanip,li2025learning,hou2024diffusion,ze20243d,wang2024equivariant,yangequibot,3d_diffuser_actor,zhu2024scaling,yao2025pick,jia2024lift3d,cao2024mamba,liu2025diffusion} have emerged as a promising direction, enabling multi-modal action generation via iterative denoising, and achieving state-of-the-art performance on a range of manipulation benchmarks.

\begin{figure}
\centering
\includegraphics[width=\linewidth]{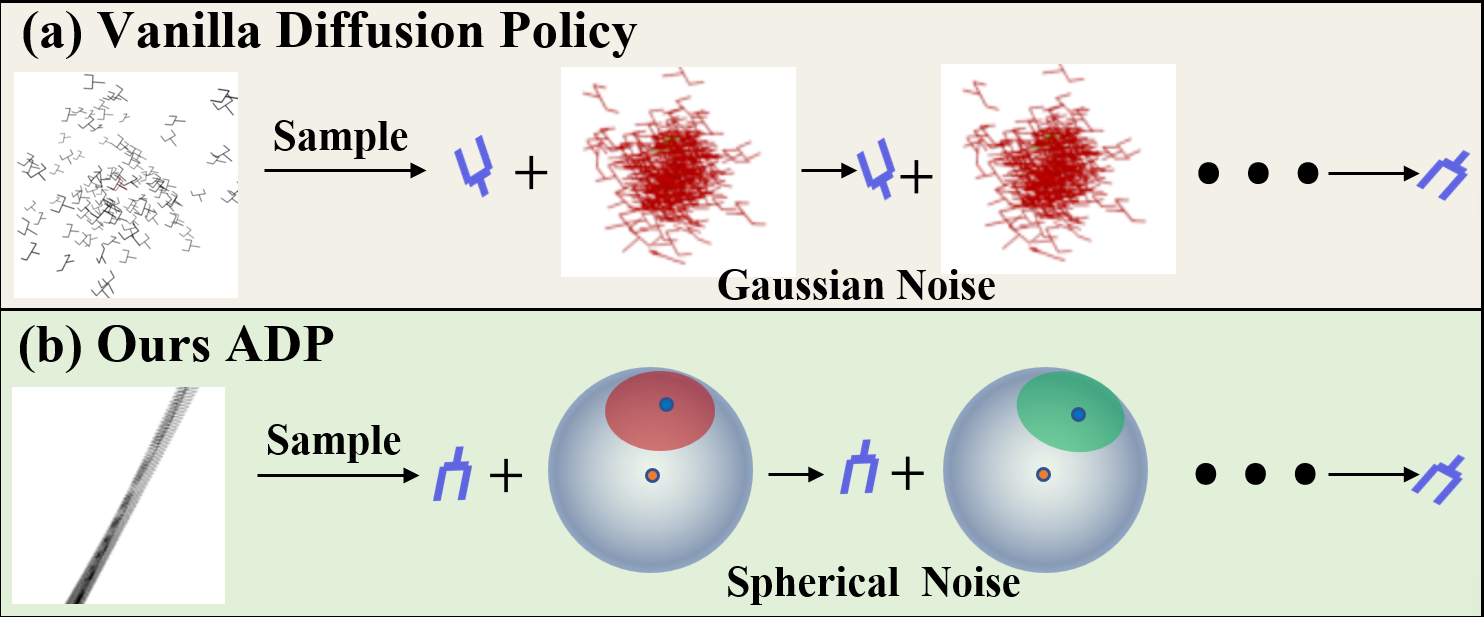}
   \caption{Comparison of (a) vanilla DPs and (b) our \textbf{ADP}. Vanilla DPs generate robot actions by progressively denoising from a random initialization, but this process typically unfolds in an unconstrained Euclidean space. In contrast, our ADP employs task-aware initialization and constrains updates along both task and spherical manifolds, yielding a more generalizable policy.}
   \label{fig:teaser}
%\vspace{-5mm}
\end{figure}

However, existing diffusion policies are typically trained offline and executed without test-time adaptation, using DDPM~\cite{ho2020denoising} or DDIM~\cite{song2020denoising} for denoising and treating action generation as an unconstrained stochastic process (see Fig.~\ref{fig:teaser}.a).
This overlooks key \textit{a priori knowledge} and \textit{task-specific structure} available at test time. For instance, the relative pose between the robot and a target object naturally suggests a direction for refinement, while coarse geometric alignment could provide better starting points. Furthermore, standard isotropic Gaussian priors inject noise uniformly across the entire action space, forcing the policy to recover from an uninformative initialization that ignores the specific characteristics of test data, resulting in inefficiencies and increased failure rates.

In this paper, we propose incorporating task-specific constraints and \textit{a priori} knowledge at test time, enabling a learned diffusion policy to dynamically adapt to task structure without retraining. As illustrated in Fig.~\ref{fig:teaser}.b, guiding the denoising process along known geometric or control-relevant manifolds (e.g., object-relative poses, contact trajectories) helps the model avoid implausible regions of the action space, resulting in more efficient inference and improved success rates. Furthermore, this structured adaptation enables the same policy to generalize across diverse tasks and environments, since it can tailor its behavior to new contexts through principled constraints rather than memorized behavior. In effect, test-time adaptation transforms a general-purpose policy into a task-specialized one on-the-fly, improving both robustness and sample efficiency.

Specifically, we introduce \textbf{ADP}, a \textit{test-time adaptive diffusion policy} that injects structure into the denoising process through three complementary mechanisms:  
%\begin{enumerate}
\begin{itemize}[leftmargin=20pt]
    \item We embed a \textit{manifold constraint} guided by the \textit{relative pose} between end-effector and target object, providing gradient-like guidance that steers the denoising trajectory toward task-relevant regions of the action space.
    \item We refine the reverse diffusion dynamics using a \textit{Gaussian spherical prior}, constraining the denoising steps to lie on a hypersphere that captures high-confidence regions of the noise distribution, reducing backtracking and improving action sampling efficiency.
    \item We propose a \textit{coarse initialization strategy} based on \textit{Fast Global Registration} between test-time point clouds of gripper and target scene, ensuring the diffusion process begins from a geometrically meaningful configuration.
\end{itemize}

ADP is \textit{training-free} and \textit{plug-and-play}, enabling seamless deployment with existing diffusion policies. Building on the pre-trained 3D Diffuser Actor~\cite{3d_diffuser_actor}, we developed \textbf{ADPro}, which improves \textit{sample efficiency}, \textit{task success}, and \textit{zero-shot generalization}, outperforming state-of-the-art diffusion-based baselines with up to \textbf{25\% fewer inference steps} and \textbf{9\% points improvement} in success rates on long-horizon tasks.

\section{Related Work}

\paragraph{Diffusion Policies for Robot Control}

Diffusion policies~\cite{chi2023diffusion,wei2025grasp,he2025learning,wangadamanip,li2025learning,hou2024diffusion,ze20243d,wang2024equivariant,yangequibot,3d_diffuser_actor,zhu2024scaling,yao2025pick,jia2024lift3d,cao2024mamba}, a new type of robotic manipulation methods that combine diffusion models~\cite{ho2020denoising,song2020denoising,song2020score,li2023dpm} with imitation learning, have attracted widespread attention for their ability to generate complex and realistic trajectories. These models are trained via behavior cloning to produce action sequences conditioned by visual inputs, robot states, or even language instructions. Existing approaches can be broadly categorized into two main types based on the input data type. The first one relies on 2D observations~\cite{chi2023diffusion,he2025learning,hou2024diffusion,wang2024equivariant,zhu2024scaling}, typically using single or multiple RGB images. Since 2D images are heavily influenced by perspective and texture, such methods tend to be sensitive to viewpoint changes and variations in object color and texture, resulting in limited robustness.  The second category leverages 3D visual information~\cite{wei2025grasp,wangadamanip,ze20243d,yangequibot,3d_diffuser_actor,yao2025pick,jia2024lift3d,cao2024mamba}, such as point clouds or RGB-D images, offering richer environmental representations and greater generalization to variations in perspective and appearance.
However, these methods face two key challenges: (1) \textbf{Limited generalization to object shapes, positions, and tasks}. Most approaches rely solely on gradient estimation from the diffusion model, without incorporating guidance from the target object to inform action updates. (2) \textbf{Slow inference}. The reverse diffusion process typically requires around 100 iterations, limiting the efficiency of action generation. Here, we address both challenges effectively by leveraging guidance from the testing data and applying spherical manifold and initial action constraints.

\paragraph{Robot Policy Generalization} In robot policy learning, the achievement of generalizable policies has been a long-standing goal~\cite{kaelbling2020foundation,li2025robotic}. Two main approaches are commonly used to improve controller generalization: training models on large, diverse datasets, and leveraging the broad generalization capabilities of large vision-language foundation models. For example, some researchers have expanded their training datasets by including more object instances or categories~\cite{finn2017deep,levine2018learning,depierre2018jacquard,young2021visual,stone2023open,wangadamanip}, novel object combinations~\cite{dasari2021transformers,jang2022bc}, diverse language instructions~\cite{jiang2022vima,nair2022learning,wei2025grasp,ren2025surfer}, and multiple robot embodiments~\cite{vuong2023open}.
Given the strong representational and generalization capabilities of vision-language models (VLMs), recent works such as RT-2~\cite{brohan2023rt}, RoboFlamingo~\cite{li2023vision}, $\pi$0~\cite{black2410pi0}, ECoT~\cite{zawalskirobotic}, and Robouniview~\cite{liu2024robouniview} have adopted pre-trained VLMs as the backbone of robotic policies to enhance robustness and generalization.  Other research explores alternative perspectives, including domain adaptation~\cite{bousmalis2018using,fang2018multi,jeong2020self,ma2024sim,wu2025momanipvla,wu2024afforddp}, continuous learning~\cite{yang2023continual,wan2024lotus,wangsparse}, and the use of prior constraints~\cite{ma2024generalizing,huangrekep,yao2025pick}.   
Despite recent advances, most methods focus on improving policy learning during training, while underutilizing the potential guidance from test-time data during inference. \textbf{A key challenge lies in how to effectively construct guidance using observational data to improve the generalization}, a problem that remains largely unexplored, and which we aim to address in this work.

\vspace{-2mm}
\section{Method}\label{sec:Method}
\begin{figure*}
\centering
\includegraphics[width=1.0\textwidth]{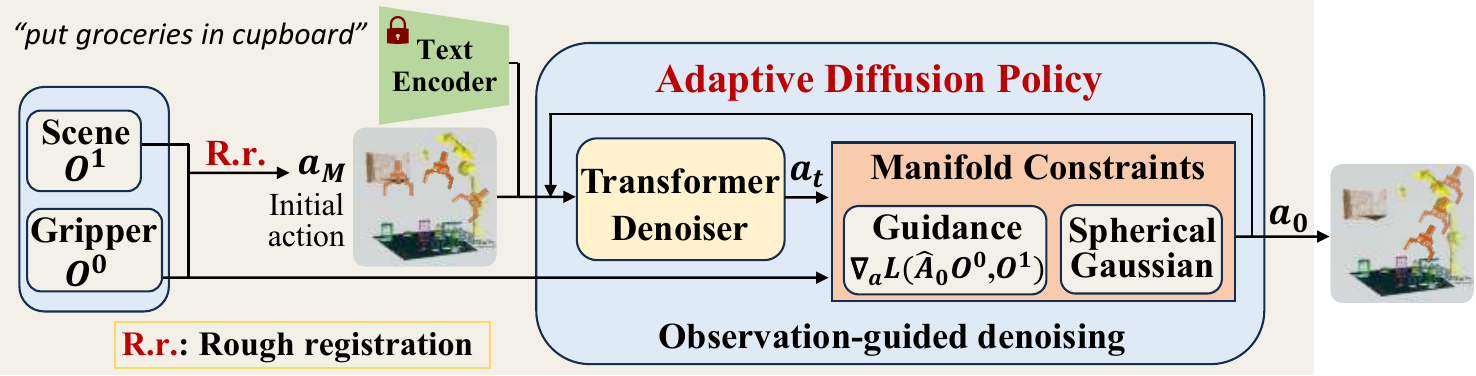}
\caption{The pipeline of our adaptive diffusion policy \textbf{ADPro}. We first propose reasonable initial noisy actions $\as_{M}$ with the rough registration module, then, denoise $\as_{M}$ under the guidance of observations and manifold constraint~(see Eq.~\eqref{eq:fdpmin} for details).} %$\ac_k$ into a vector $\as_k$
\label{fig:pipeline}
%\vspace{-3mm}
\end{figure*}

%\vspace{-2mm}
Action generation inherently involves producing non-Euclidean data, as actions reside within the $\SE(3)$ Lie group. Therefore, integrating a more efficient diffusion policy with a well-designed initial noisy action proposal can greatly enhance generation efficiency. As illustrated in Fig.~\ref{fig:pipeline}, the proposed adaptive diffusion policy for manipulation action generation accelerates the process through two key improvements. First, a manifold-constrained denoising – guiding updates along both task and spherical manifolds, effectively enhancing its generalizability and preventing unnecessary backtracking~(see Fig.~\ref{fig:rlbenchcomparison_astep}). Second, task-aware initialization – driving the diffusion process towards a more accurate direction instead of random noise, effectively reducing the number of denoising steps required. Together, these components transform action generation from blind stochastic sampling into a geometry-aware, adaptive optimization process.

\subsection{Problem Statement.}\label{sec:setup}

Given a set of expert demonstrations containing continuous-valued action-observation pairs, denoted as $\mathcal{D} = \{(\ob^k,\ac^k )\}_{k=1}^N$, our objective is to fit a visuomotor policy that maps the observation space $\ob$ to the action space $\ac$. In this work, each observation $\ob$ is two posed RGB-D images or point clouds. Each action $\ac := (\mathbf{v}, \mathbf{R}, w)$ is a 7D pose, defined by position $(x,y,z)$, rotation $(roll,pitch,yaw)$, and the gripper width $w$. Consequently, the pose $(\mathbf{v}, \mathbf{R})$ of action lies on an $\SE(3)$ manifold, where $\mathbf{R} \in \SO(3)$ is the rotation matrix, and $\mathbf{v} \in \mathbb{R}^3$ is the translation vector. 

Let $\mathbf{P}^k\in\mathbb{R}^{4\times 4}$ be the current position of the gripper in the world frame, and action $\ac^k$ specifies a desired pose of the gripper. The pose can be either absolute ($\mathbf{P}^{k}=(\mathbf{R}^k, \mathbf{v}^k)$, also called position control) or relative ($\mathbf{P}^{k}=(\prod_{i=1}^k\mathbf{R}^i +\sum_{i=1}^k\mathbf{v}^i$), also called velocity control). To facilitate noise injection and removal using addition and subtraction, as in the standard diffusion process, we vectorize $\ac^k$ into a vector $\as^k$ during diffusion and denoising. After denoising, the noise-free action vectors are converted back into $\mathbf{R}$, $\mathbf{v}$, and $w$ for manipulation. 

For trajectory generation, $\ob = \{\ob^{k-(m-1)},\dots, \ob^{k-1},\ob^k\}$ , $\as=\{\as^k, \as^{k+1}, \dots, \as^{k+(n-1)} \}$ where $m$ is the number of history steps observed and $n$ is the number of future action steps.  
%Given that diffusion policies involve both multiple-action trajectories across different positions and a single denoising trajectory for each action at a given position, to avoid ambiguity, we follow the common notation conventions and use a \textbf{superscript} to denote individual action $\as^k$ within action trajectories and a \textbf{subscript} to represent noisy action trajectory $\as_t$ during the denoising process for a single action.
Since diffusion policies involve both multi-step action trajectories across time and a denoising trajectory for each individual action, we follow standard notation conventions to avoid ambiguity: \textbf{superscripts} denote individual actions $\as^k$ within a manipulation trajectory, while \textbf{subscripts} denote the noisy action trajectory $\as_t$ during the denoising process of diffusion for a single action.

\subsection{Preliminaries}
\paragraph{Diffusion Policy} Diffusion policy~(DP)~\cite{chi2023diffusion} formulates the manipulation process as a Markov process and leverages DDPM~\cite{ho2020denoising} to model the manipulation policy with multimodal visuomotor inputs $\ob$. It learns a noise prediction function $\varepsilon_\theta(\ob, \as+\varepsilon_t, t)=\varepsilon_t$ using a network $\varepsilon_\theta$ parameterized by $\theta$. 
To train the network $\varepsilon_\theta$ to accurately predict the noise component of the input $\as+\varepsilon_t$, random noise $\varepsilon_t$, conditioned on a randomly sampled denoising step $t$ is added to $\as$. Then, DP minimize the loss $\mathcal{L}=||\varepsilon_\theta(\ob, \as + \varepsilon_t, t) - \varepsilon_t||^2$, which is
equivalent to matching the denoising scores $\nabla_{\as} \log p(\as|\ob)$~\cite{chi2023diffusion}. During inference, given observation $\ob_k$, DP performs a sequence of $T$ denoising steps starting from a random action $\as_{T}^k\sim \mathcal{N}(0, I)$ to generate an action $\as_0^k$ defined by
\begin{equation}
\label{eqn:ddpm}
    \as_{t-1}^k=\alpha_t(\as_t^k - \gamma_t \varepsilon_\theta(\ob^k, \as_t^k, t) + \epsilon),
\end{equation}
where $\epsilon\sim \mathcal{N}(0, \sigma_t^2I)$. $\alpha_t, \gamma_t, \sigma_t$ are functions of the denoising step $t$ (also known as the noise schedule). The action $\as_0$ is expected to be the corresponding action from the expert policy $\pi: \ob \mapsto \as$. 

\paragraph{Training-Free Guided Diffusion Models} 
Classifier guidance~\cite{classifier} is the first work that utilizes the pre-trained diffusion model for conditional image generation in a training-free fashion. Specifically, considering the Bayes rule $p(x|y)=p(y|x)p(x)/p(y)$, it introduces the given condition with an additional likelihood term $p(x_t | y)$:
\begin{equation}
    \nabla_{x_t} \log p(x_t | y) = \nabla_{x_t} \log p(x_t) + \nabla_{x_t} \log p(y | x_t), 
\end{equation}
where $y$ is the class label, $x_t \sim \mathcal{N}(\sqrt{\bar{\alpha}_t} x_0, (1-\bar{\alpha}_t) I)$, $\bar{\alpha}_t:=\prod_{s=1}^t \alpha_s$. Without training the time-dependent classifier to estimate $p(y|x_t)$, these training-free guided diffusion models~\cite{classifier,chungdiffusion,yang2024guidance} just need a pre-trained diffusion prior $\varepsilon_\theta$ and a differentiable loss $L(x_0,y)$ which is defined on the support of $x_0$. They use Tweedie’s formula to calculate $\nabla_{x_t} \log p(y | x_t)$ by estimating $\hat{x}_0$ based on $x_t$:
\begin{equation}
\begin{split}
    \hat{x}_0(x_t) \approx \mathbb{E}[x_0 | x_t] &= (x_t + \sqrt{1-\bar{\alpha}_t} \nabla_{x_t} \log p(x_t)) / \sqrt{\bar{\alpha}_t},\\
    \nabla_{x_t} \log p(y | x_t) \approx & \nabla_{x_t} \log p(y | \hat{x}_0(x_t)) = \eta \nabla_{x_t} L(\hat{x}_0(x_t),y).  
\end{split}
\end{equation}
Then they use the estimated likelihood of $L(x_0,y)$ for additional correction step:
\begin{equation}
    x_{t-1} = \underbrace{DDIM(x_t, \varepsilon_\theta(x_t,t),t)}_{\text{sampling step}} - \underbrace{\eta \nabla_{x_t} L(\hat{x}_0(x_t),y)}_{\text{correction step}}.
\end{equation}

\subsection{Manifold-Constrained Denoising}
%Guided diffusion models have achieved significant success in generating Euclidean data, such as images. Given the limitations of diffusion policies in generalizing across different tasks, we believe that training-free guided diffusion policies offer an effective solution. However, existing guided diffusion models primarily focus on Euclidean data generation, and in the domain of robotic manipulation, constructing effective guidance remains a challenge.

Guided diffusion models have shown strong generalization at inference, achieving remarkable success in generating Euclidean data such as images. Yet, most existing approaches remain limited to image domains, and in robotic manipulation, constructing effective guidance remains a key challenge.

In robotic manipulation, vanilla DPs (non-guided) adopt $\varepsilon_\theta (\ob^k, \as^k_t, t)$ as the denoising direction, which serves as the primary update for action refinement. Yet, this approach does not explicitly exploit the target object to optimize actions more effectively. To address this, we introduce ADP, which incorporates geometry-aware updates. Specifically, we identify two complementary manifolds that constrain and guide the denoising process: the task manifold and the spherical manifold.

\paragraph{Task manifold} Robot actions lie on the $\SE(3)$ Lie group, where the relative pose between the end-effector and the target object defines a natural gradient direction toward task success. To exploit this structure, we incorporate observation guidance $\nabla_{\as}L(\cdot,\cdot)$ into the diffusion process, steering updates along the geodesic paths of the task manifold. This yields the following guided variant of the diffusion policy:
\begin{align}
\Tilde{\as}^k_{t-1} &= DDPM(\as_t^k, \varepsilon_\theta(\ob^k, \as_t^k, t),t) \label{eq:dp}\,, \\
\as^k_{t-1} &=\Tilde{\as}^k_{t-1} - \eta \nabla_{\as} L(\Tilde{\ac}^k_{t-1} \ob^{k,0}, \ob^{k,1}) \label{eq:dpmin}\,.
\end{align}
Here, $\ob^{k,0}$ and $\ob^{k,1}$ represent the gripper and scene point clouds extracted from $\ob_k$, respectively. $\Tilde{\ac}^k_{t-1}$ denotes the matrix form of $\Tilde{\as}^k_{t-1}$, while $L(\cdot,\cdot)$ is Chamfer distance. 

However, since $\Tilde{\ac}^k_{t-1}$ is a noisy action rather than the final action executed by the robot, the guidance provided by Eq.~\eqref{eq:dpmin} is not sufficiently accurate. Inspired by DDPM~\cite{ho2020denoising}, we obtain the noise-free action $\hat{\as}^k_{0}$ from $\Tilde{\as}^k_{t-1}$ using the diffusion function, and reformulate Eq.~\eqref{eq:dpmin} as 
\begin{align}
&\hat{\as}^k_{0} =(\Tilde{\as}^k_{t-1} -\sqrt{1-\bar{\alpha}_t}\varepsilon_\theta(\ob^k, \as_t^k, t))/\sqrt{\bar{\alpha}_t}\,,\label{eq:adp}\\
&\as^k_{t-1} =\Tilde{\as}^k_{t-1} - \eta \nabla_{\as} L(\hat{\ac}^k_{0} \ob^{k,0}, \ob^{k,1}) \,,\label{eq:adptem}
\end{align}
where $\hat{\ac}^k_{0}$ denotes the matrix form of $\hat{\as}^k_{0}$,

With the observation guidance $\nabla_{\as}L(\cdot,\cdot)$, Eq.~\eqref{eq:adptem} constrains the update direction of the inverse diffusion process and shortens the diffusion trajectory. Moreover, by incorporating observation point clouds from the test phase into the guidance, Eq.~\eqref{eq:adptem} effectively enhances the generalization of DPs. However, it implicitly relies on a strong assumption—the linear manifold, which may lead to manifold deviation and unnecessary backtracking in the trajectory.

\begin{figure*}[tbp!]
%\vspace{-5mm}
\centering
\subfloat{\includegraphics[width=0.9\textwidth]{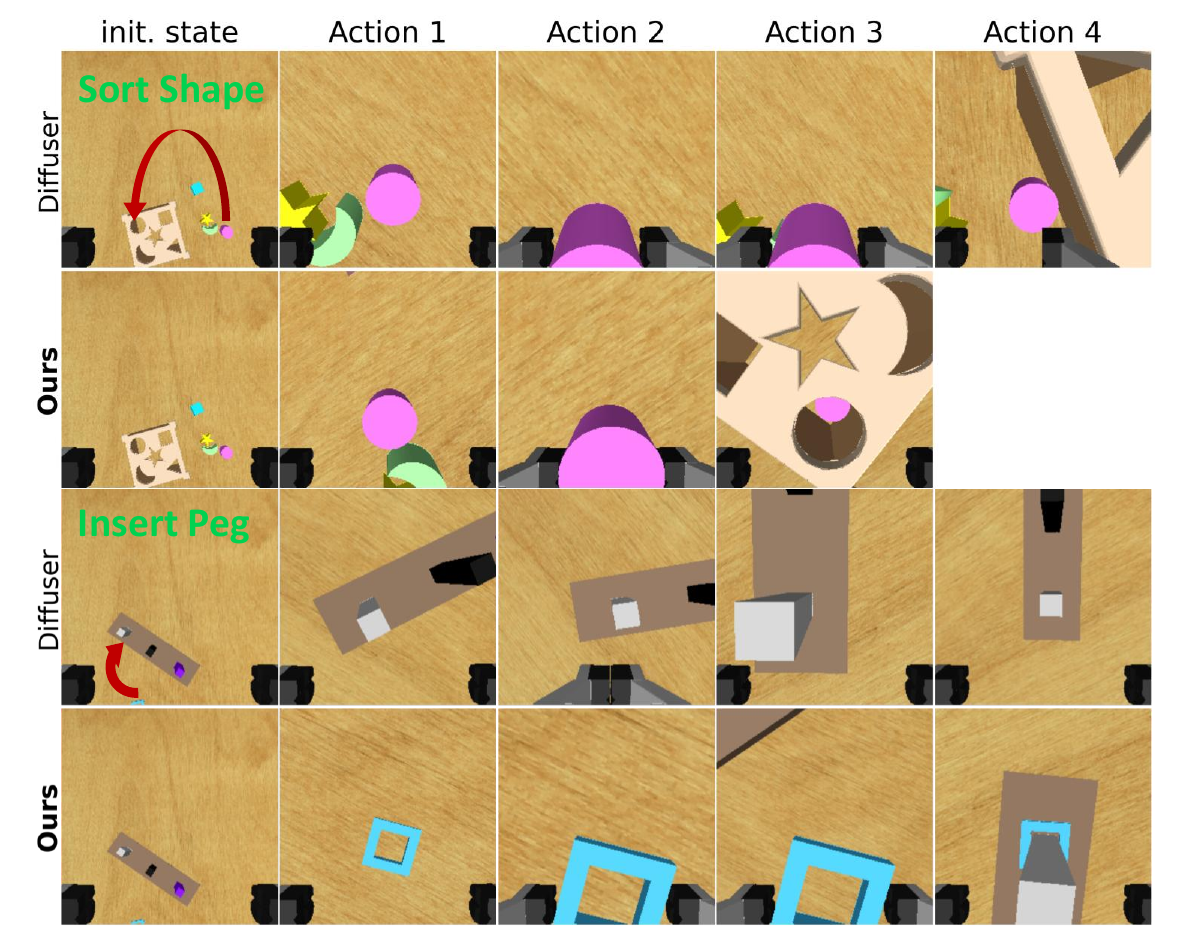}}
\vspace{-3mm}
\caption{Trajectory comparison between the vanilla diffusion policy and our \textbf{ADPro}. We illustrate two distinct tasks:  `\textit{sort shape}' and  `\textit{insert peg}'. For each task, we visualize the initial state followed by successive observations captured from the wrist-mounted camera after each action. ADPro completes the first task with only three actions, demonstrating both efficiency and accuracy.} 
\label{fig:visulization}
\end{figure*}

\paragraph{Spherical manifold} To minimize unnecessary backtracking in the diffusion process~(illustrated as Diffuser in Fig.~\ref{fig:rlbenchcomparison_astep}), we introduce a manifold constraint into the reverse diffusion policy, ensuring that the denoising updates align more closely with the manifold's geodesic path. Specifically, we adopt the Gaussian spherical prior~\cite{yang2024guidance} to refine the reverse diffusion step size. The Gaussian spherical prior constrains the noise sampling space to a uniform distribution on the hypersphere of radius $\sqrt{d}\sigma$, keeping updated step size within high-confidence intervals. Here, $d$ is the product of the action dimension, the batch size, and the number of actions $n$ to be predicted. As remarked in \cite{yang2024guidance}, when $d$ is sufficiently large, a $d$-dimensional isotropic Gaussian distribution $\mathcal{N}(0, \sigma^2I)$ is close to a uniform distribution on the hypersphere of radius $\sqrt{d}\sigma$. Then, the refined training-free adaptive diffusion policy is formulated as:
\begin{equation}
\label{eq:fdpmin}
\as^k_{t-1} =\Tilde{\as}^k_{t-1} - \sqrt{d}\sigma \frac{\nabla_{\as} L(\hat{\ac}^k_{0} \ob^{k,0}, \ob^{k,1})}{||\nabla_{\as} L(\hat{\ac}^k_{0} \ob^{k,0}, \ob^{k,1})||} \,,
\end{equation}

By unifying these constraints under the same principle of manifold-constrained denoising, ADP transforms the reverse diffusion process into a structured trajectory: task manifolds provide directionality, spherical manifolds offer stability, and together they yield more efficient and robust action generation.

\subsection{Task-Aware Initialization}
Given the end-effector $\ob^{k,0}$ and a target scene $\ob^{k,1}$, the search space for initial actions should be restricted to a specific local region rather than spanning the entire noise space, as is common in existing diffusion policies. Relying on a predefined noise distribution for sampling across the full noise space not only increases the complexity of subsequent denoising but also lowers the manipulation success rate. A more effective way is to propose a reasonable initial action by computing a coarse registration, which simplifies denoising and reduces the number of diffusion steps needed. 

To this end, ADP introduces a task-aware initialization module. Rather than sampling from an uninformed prior, we compute a coarse alignment between the gripper and the target scene point clouds. In our implementation, we employ the Fast Global Registration (FGR) algorithm~\cite{zhou2016fast} to align the observed data $\ob^{k,0}$ with $\ob^{k,1}$. FGR outputs a rotation matrix $\mathbf{R} \in \SO(3)$ and a translation vector $\mathbf{v} \in \mathbb{R}^3$, which together define the initial noisy action $\as^k_{M}$, corresponding to the noisy output at step $M$ ($M \leq T$) of the original diffusion process. Compared with the classical Iterative Closest Point (ICP) algorithm~\cite{besl1992method}, FGR offers greater efficiency and robustness, particularly in handling noisy or large-scale point clouds. 
\begin{equation}
\label{eq:adpinitial}
\as^k_{M} = FGR(\ob^{k,0}, \ob^{k,1})
\end{equation}

In practice, we randomly sample 4096 points from each point cloud and use FGR to align the end-effector point cloud to the scene point cloud. A maximum of 6–10 iterations is sufficient, as only a coarse estimate is required. By initializing the diffusion process in a task-relevant region, we effectively reduce the search space and accelerate convergence. Ablation studies confirm that removing this component significantly degrades both success rate and efficiency.

\subsection{Unified Adaptive Diffusion Policy and Error Analysis} \label{sec:Manipulation}
Bringing these components together, ADP functions as a test-time optimizer for pretrained diffusion policies: \textbf{(i)} initialization places the process near a plausible region, \textbf{(ii)} manifold constraints guide updates along geometry rather than noise, and \textbf{(iii)} the spherical constraint regularizes step magnitudes to prevent instability. In robotic manipulation, ADP generates control actions and integrates seamlessly with existing diffusion policies in a plug-and-play manner, without retraining. Concretely, we first generate noisy initial actions $\as^k_{M}$ using Eq.~\eqref{eq:adpinitial}, and then iteratively denoise them via Eq.~\eqref{eq:dp}, Eq.~\eqref{eq:adp}, and Eq.~\eqref{eq:fdpmin} until the final noiseless action $\as^k_{0}$ is obtained.

\begin{table*}[t]
    \centering
    %\footnotesize
    %\small
    %\vspace{-5mm}
    \caption{The mean success rate on RLBench across three random seeds. ADPro outperforms all prior arts on most tasks by a large margin. \textbf{Bold} indicates the best; \underline{underlining} shows the second best.}
    %\vspace{-2mm}
    \begin{adjustbox}{width=1.0\textwidth}
    \tb{@{}l|c|c|c|c|c|c|c|c|c|c|c|c|c|c|c|c|c|c|c@{}}{1.0}{
    \toprule
    & \cellcolor{ImportantColor}Avg. & open & slide & sweep to & put in & turn & put in & place & stack & sort &insert& drag& place& close & meat off &stack & push &screw &put in  \\
    & \cellcolor{ImportantColor}Success $\uparrow$ & drawer & block & dustpan & safe & tap & drawer & wine & cups & shape & peg& stick & cups& jar & grill &blocks &buttons &bulb &cupboard \\
    \midrule
    C2F-ARM & \cellcolor{ImportantColor}20.1 &  20 & 16 & 0 & 12 & 68 & 4 & 8 & 0 & 8 & 4&24 &24 &0 &20 &0  &72 &8 &0 \\
    PerAct & \cellcolor{ImportantColor}49.4 & 88.0 & 74.0& 52.0 & 86.0 & 88.0 & 51.2 & 44.8 & 2.4 & 16.8 & 5.6&89.6 &2.4 &55.2 &70.4 &26.4 &0  &17.6 &28.0 \\
    HiveFormer & \cellcolor{ImportantColor}45 & 52 & 64 & 28 & 76 & 80 & 68 & 80 & 0 & 8 & 0& 76 &0  &52  &\textbf{100} &8  &84 &8 &32\\
    PolarNet & \cellcolor{ImportantColor}46 & 84 & 56 & 52 & 84 & 80 & 32 & 40 & 8 & 12 & 5& 92&0  &36 &\textbf{100} &4  &96 &44 &12\\
    RVT & \cellcolor{ImportantColor}62.9 & 71.2 & 81.6 & 72.0 & 91.2 & 93.6 & 88.0 & 91.6 & 26.4 & 36.0 & 11.2& 99.2&4.0  &52.0 & 88.0 &28.8 &\textbf{100}  &48.0 &49.6\\
    Act3D & \cellcolor{ImportantColor}63.2 & 78.4 & 96.0 & \underline{86.4} & 95.2  & 94.4  & 91.2 & 59.2  & 9.6  & 29.6 & 24.0& 80.8&3.2 & \textbf{96.8} & 95.2 &4.0 &93.6 &32.8 &67.2 \\
    Diffuser &  \cellcolor{ImportantColor}81.3 & \underline{89.6} & \underline{97.6} & 84.0 & \underline{97.6} & \underline{99.2} & \underline{95.0} & \underline{93.6} & 47.2 & \underline{44.0} & 65.6& \textbf{100.0}&24.0 &96.0 & 96.8 & \underline{66.8}  &98.4 &\underline{82.4} &\underline{85.6} \\
    VPDD &\cellcolor{ImportantColor}56.2& - & 70.7  & 40.0 & 70.7 &88.7  & 24.0 &60.0 & \underline{56.0} & - & \underline{73.7}& 66.7 &\textbf{30.7}  &37.3 &73.3 & 56.0  &58.7  &61.3 &30.67   \\ 
    \textbf{Ours}  &\cellcolor{ImportantColor} \textbf{83.9} & \textbf{93.1} & \textbf{100.0} & \textbf{97.0}  & \textbf{98.5}  & \textbf{99.5}  &\textbf{95.2}  & \textbf{94.0} & \textbf{56.5} & \textbf{50.0} & \textbf{74.5}&\textbf{100.0}&\underline{25.4}&\underline{96.2} & 95.6 &\textbf{67.3}  &\underline{98.6} &\textbf{83.1} &\textbf{85.9} \\
    \bottomrule
    }
    \end{adjustbox}
    \label{tab:peract}
    %\vspace{-1mm}
\end{table*}

\iffalse
\begin{figure*}
\vspace{-3mm}
\centering
\includegraphics[width=1.0\textwidth]{figures/rl_actions_plot_2.png}
\vspace{-5mm}
\caption{Comparison on components ($x$, $z$, $r_2$, $r_3$) of the full action for task `\textit{sweep to dustpan}'. Full action comparisons are available in the supplementary. The horizontal and vertical axes represent action steps and parameter values, respectively. Our ADPro effectively mitigates coordinate and angle backtracking behaviors in the vanilla diffusion policy.}
\label{fig:rlbenchcomparison_astep}
\vspace{-3mm}
\end{figure*}
\fi

\begin{figure*}[tbp!]
    \centering
    \begin{subfigure}{\textwidth}
        \centering
        \includegraphics[width=1.\textwidth]{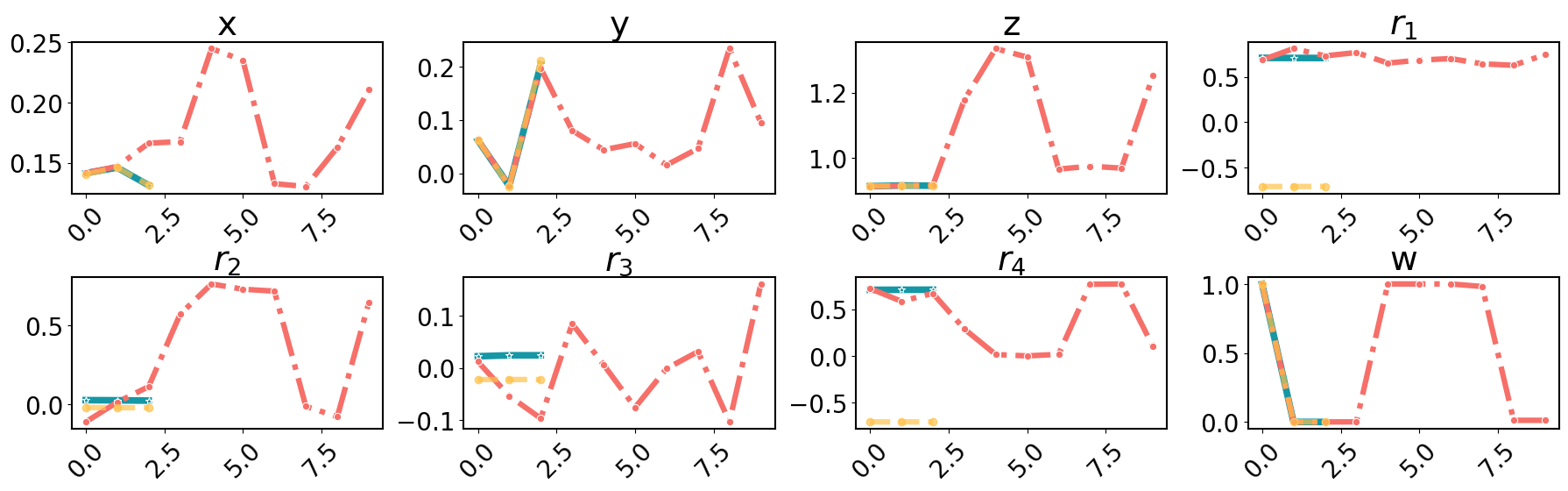}
        \caption{Open drawer}
    \end{subfigure}
    \begin{subfigure}{\textwidth}
        \centering
        \includegraphics[width=1.\textwidth]{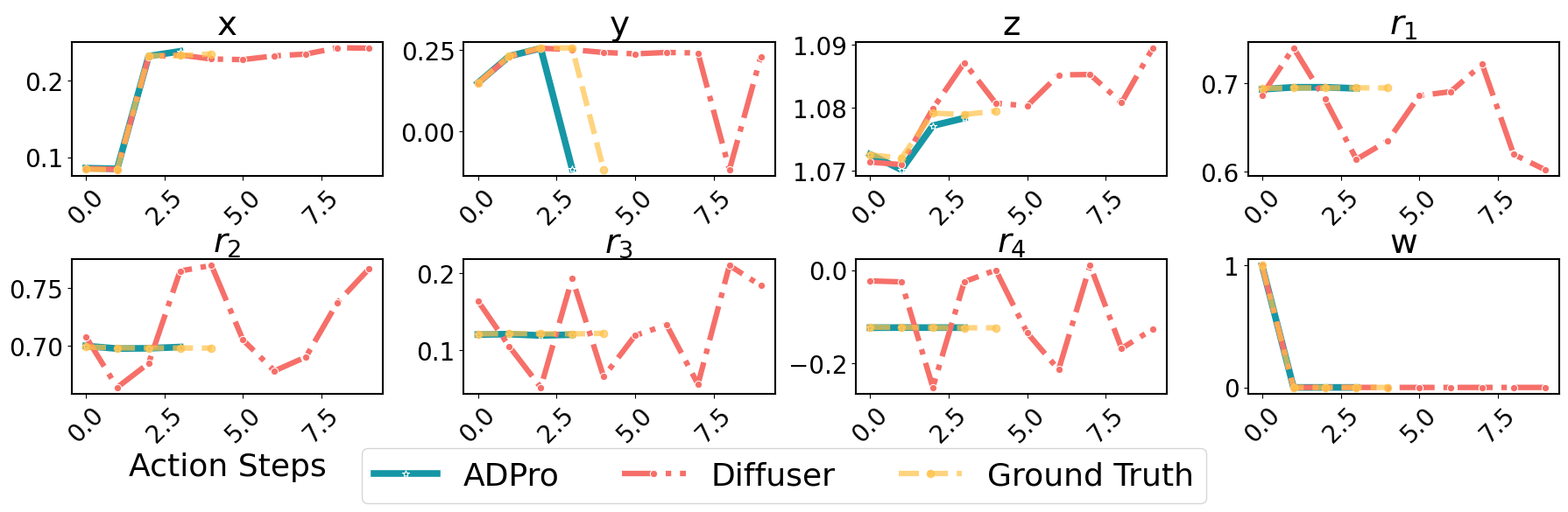}
        \caption{Sweep to dustpan}
    \end{subfigure}
\caption{Comparison on all components ($x$, $y$, $z$, $r_1$, $r_2$, $r_3$, $r_4$, $w$) of the full action for tasks `\textit{open drawer}' and `\textit{sweep to dustpan}'. The horizontal and vertical axes represent action steps and parameter values, respectively. Our ADPro effectively mitigates coordinate and angle backtracking behaviors in the vanilla diffusion policy.}
\label{fig:rlbenchcomparison_astep}
\end{figure*}

\begin{figure*}
%\vspace{-3mm}
\centering
\includegraphics[width=1.0\textwidth]{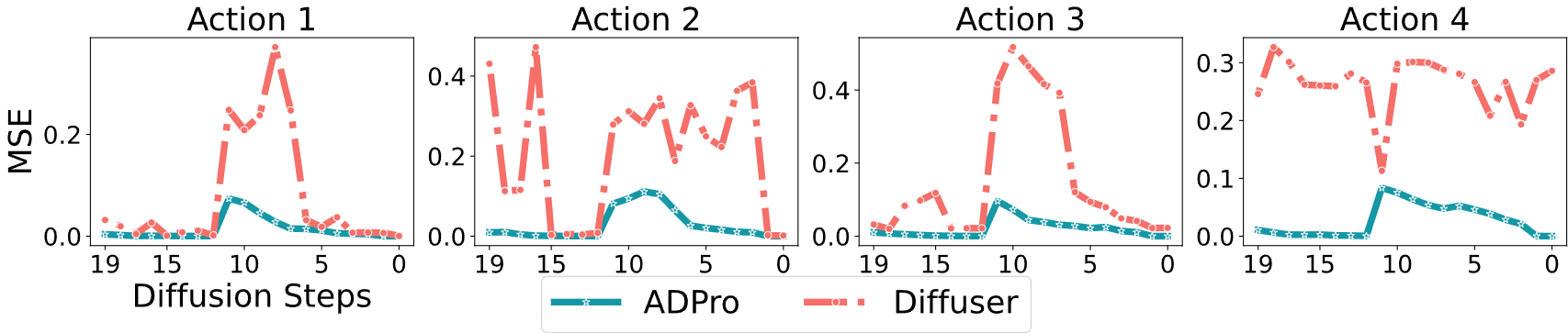}
\vspace{-3mm}
\caption{Diffusion-step evolution for the first four actions in task `\textit{sweep to dustpan}'. The MSE of Diffuser exhibits large fluctuations, indicating significant backtracking behavior in its trajectory.} 
\label{fig:rlbenchcomparison_dstep}
\vspace{-3mm}
\end{figure*}

Considering that 3D robot policies have shown superior generalization over 2D policies across varying camera viewpoints and are more effective in handling novel viewpoints during testing~\cite{3d_diffuser_actor,ze20243d,jia2024lift3d}, we implement ADP with 3D Diffuser Actor to create ADPro. ADPro’s conditional noise predictor $\varepsilon_\theta$ , leverages a pretrained 3D Transformer-based network. To further enhance the model's generalization across diverse tasks, we incorporate task instructions as textual prompts encoded by the text encoder of the vision-language model CLIP~\cite{radford2021learning}. The resulting text features are then fused with visual features through a cross-attention mechanism, enabling a more effective integration of semantic and visual information.

\textbf{Error Analysis.} The robustness of the proposed ADP depends on both the FGR algorithm and the pre-trained diffusion model. When $\as^k_M$ in Eq.~\ref{eq:adpinitial} satisfies $||\as^k_M -\as^k_0||<\delta_1$, the noisy action $\hat{\as}^k_M$ of the pretrained $\varepsilon_\theta$ at steps $M$ satisfies $||\hat{\as}^k_M-\as_k^0||<\delta_2$, then, $||\as^k_M- \hat{\as}^k_M|| <\delta_1+\delta_2$. Furthermore, $\delta_1$ is controllable and can be reduced by increasing the number of iterations of the FGR algorithm, which in turn lowers $\delta_1$ and reduces the error upper bound. This ensures the robustness of the proposed ADP.

\section{Experiments}\label{sec:experiments}
We designed experiments to answer the following questions:  \textbf{i) Does the proposed diffusion policy improve manipulation performance? ii) How does ADPro perform in terms of generalization and efficiency? iii) Can ADPro be applied to real-world robotic manipulation tasks?} 
We first demonstrate the performance improvements of our proposed ADPro on RLBench~\cite{james2020rlbench} and CALVIN~\cite{mees2022calvin}. Then, in Sections~\ref{sec:calvin} and \ref{sec:generalizability}, we validate its enhancements in generalization and efficiency on CALVIN and a real-world dataset. Finally, we conduct ablation studies in Section~\ref{sec:ablation} to verify the effectiveness of each core module.

%We first demonstrate the performance improvements of our proposed ADPro in Sections~\ref{sec:RLBench} and \ref{sec:calvin}. Then, in Sections~\ref{sec:calvin} and \ref{sec:generalizability}, we validate its enhancements in generalization and efficiency. Finally, we conduct ablation studies in Section~\ref{sec:ablation} to verify the effectiveness of each core module.

\subsection{Baselines}
We evaluate the proposed methods with thirteen baselines in four datasets, including C2F-ARM~\cite{james2022coarse}, PerAct~\cite{shridhar2023peract}, HiveFormer~\cite{guhur2023instruction}, PolarNet~\cite{chen2023polarnet}, RVT~\cite{goyal2023rvt}, Act3D~\cite{gervet2023act3d}, HULC~\cite{mees2022hulc}, RoboFlamingo~\cite{li2023vision}, SuSIE~\cite{black2023zero}, GR-1~\cite{wuunleashing}, and three DPs: 3D Diffusion Policy~(DP3)~\cite{ze20243d}, 3D Diffuser Actor~(Diffuser)~\cite{3d_diffuser_actor}, VPDD~\cite{he2025learning}. %We conduct tests using the default settings, official implementations, and pre-trained models for all baseline methods for a fair comparison.
For VPDD, we use the results reported in their official paper. Since no publicly available pretrained model of DP3 for CALVIN exists, we train it on CALVIN to ensure a fair comparison. For the remaining baselines, we adopt the results from 3D Diffuser Actor.
Firstly, we leverage benchmarks RLBench and CALVIN to present the effectiveness of the proposed \textbf{ADP} on task success rates and inference speed. Meanwhile, we demonstrate the generalization of our policy in real-world scenes with Franka Panda Robot. We also compare against ablative versions that do not consider manifold or initial noise constraints.

% We map the language task instruction to language tokens using a pre-trained CLIP language encoder~\cite{radford2021learning}

\subsection{Performance on RLBench}\label{sec:RLBench}
RLBench is a simulation environment where a Franka Panda Robot is used to manipulate the scene with BiRRT~\cite{kuffner2000rrt} as its motion planner. On RLBench, all methods need to predict the next end-effector keypose. We evaluate ADPro on 18 tasks as ~\cite{3d_diffuser_actor}, each with 2-60 variations. We evaluate policies by task completion success rate, the ratio of execution trajectories that achieve the goal conditions specified in the language instructions~\cite{gervet2023act3d,shridhar2023peract}. Tab.~\ref{tab:peract} presents the evaluation metrics. The baseline performances are reported from Diffuser, and our results are averaged over three random seeds.

As shown in Table~\ref{tab:peract}, our method consistently outperforms the vanilla diffusion policy (Diffuser), with particularly notable gains on the tasks `\textit{sweep to dustpan}', `\textit{sort shape}', and `\textit{insert peg}', achieving an average success rate improvement of 2.6 points. Figure~\ref{fig:visulization} further illustrates these advantages: on `\textit{sort shape}', ADPro completes the task in three direct actions, whereas Diffuser requires four with detours; on `\textit{insert peg}', ADPro achieves early alignment and inserts without backtracking, while Diffuser needs an extra alignment action to finish the task. This visual comparison reinforces the efficiency and stability benefits of our approach. Furthermore, Fig.~\ref{fig:rlbenchcomparison_astep} compares all parameters of a 7-DOF robotic arm—including position ($x$, $y$, $z$), rotation ($r_1$, $r_2$, $r_3$, $r_4$), and gripper width ($w$)—across tasks. Our method converges faster and more accurately to the target grasp configuration, while Diffuser exhibits significant backtracking. These results validate the effectiveness of our guidance mechanism and initialization strategy. Notably, ADPro completes tasks in fewer steps, demonstrating superior efficiency in generating high-quality manipulation trajectories.

%As shown in Fig.~\ref{fig:visulization}, on `\textit{sort shape}' ADPro completes the task in three actions with a direct approach, while Diffuser needs four and detours; on `\textit{insert peg}' ADPro aligns early and inserts without backtracking, unlike Diffuser’s corrective step. This visual comparison echoes our gains in efficiency and stability. Additionally, Fig.~\ref{fig:rlbenchcomparison_astep} illustrates the evaluation of all parameters of a 7-DOF robotic arm, including position ($x$, $y$, $z$), rotation ($r_1$, $r_2$, $r_3$, $r_4$), and the gripper width ($w$), during an episode for different tasks. As observed in these figures, our method exhibits a faster and more accurate convergence toward the target grasping configuration compared to Diffuser. This demonstrates the effectiveness of our proposed guidance mechanism and initialization strategy. Notably, ADPro also completes tasks in fewer steps, further underscoring its superior efficiency in planning high-quality manipulation trajectories. 

\begin{figure}
\vspace{-1mm}
\centering
\includegraphics[width=1.0\linewidth]{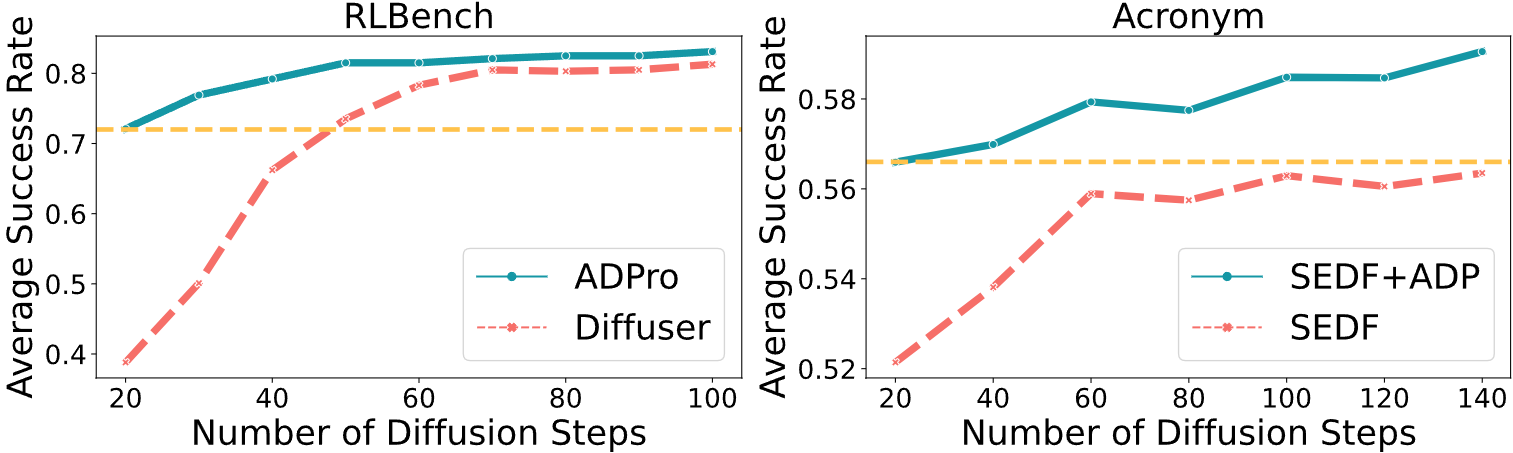}
\vspace{-3mm}
\caption{Evaluating the influence of diffusion steps on two benchmarks: RLBench and Acronym.} 
\label{fig:combined_dsteps}
\vspace{-2mm}
\end{figure}

In Fig.~\ref{fig:rlbenchcomparison_dstep}, we examine the inverse diffusion process for each action prediction. To enhance readability, we compute the Mean Squared Error (MSE) between the predicted actions and the ground truth, where values closer to zero indicate better predictions. The results demonstrate that incorporating our guidance during generation substantially reduces unnecessary backtracking and improves stability, yielding a much smoother generation trajectory. Notably, for `\textit{Action 2}' and `\textit{Action 4}', due to alignment, the initial position of the diffusion process is closer to the ground truth, which facilitates faster convergence toward an accurate prediction and reduces the number of diffusion steps required. In the left panel of Fig.~\ref{fig:combined_dsteps}, we vary the number of diffusion steps used for generation and evaluate the average success rate on RLBench. We find that our method significantly outperforms the baseline, particularly when using a small number of diffusion steps. Notably, our method achieves a $72\%$ success rate with only 20 steps, whereas the baseline requires at least 50 steps to reach comparable performance. Moreover, our method consistently improves upon the baseline across all diffusion step settings.

\begin{table}[t]
\caption{Zero-shot long-horizon evaluation on CALVIN.}
\footnotesize
\renewcommand{\tabcolsep}{1.3mm}
\centering
%\vspace{-2mm}
\begin{tabular}{@{}l|ccccc|c|c@{}}
\toprule
\multirow{2}{*}{Method}  & \multicolumn{5}{c|}{Task completed in a row}& Avg. & Avg. Time(s) \\
& 1 & 2 & 3 & 4 & 5&Len $\uparrow$& / NDS $\downarrow$\\
    \midrule
    HULC  & 41.8 & 16.5 & 5.7 & 1.9 & 1.1 & \cellcolor{ImportantColor}0.67& \\
    RoboFlamingo  & 82.4 & 61.9 & 46.6 & 33.1 & 23.5 & \cellcolor{ImportantColor}2.48& \\% failed because can load openGL, try on a local machine with graphic interface maybe work
    SuSIE  & 87.0 & 69.0 & 49.0 & 38.0 & 26.0 &\cellcolor{ImportantColor} 2.69& \\% No module named 'jaxrl_m' can be found
    GR-1 & 85.4 & 71.2 & 59.6 & 49.7 & 40.1 &\cellcolor{ImportantColor} 3.06& \\
    \hline
    DP3  & 53.9 & 44.7 & 38.0 & 34.3 & 29.0 & \cellcolor{ImportantColor}2.00 &36.1 / 25 \\
    \textbf{DP3+ADP} & 78.4 & 63.7 & 54.9 & 39.8 & 30.2 & \cellcolor{ImportantColor}2.67 &27.3 / 10 \\
    \hline
    Diffuser & 93.8 & 80.3 & 66.2 & 53.3 & 41.2 & \cellcolor{ImportantColor}3.35&48.3 / 25 \\
    \textbf{Ours} &  \textbf{94.7}  & \textbf{83.0} & \textbf{73.6} & \textbf{61.4} & \textbf{51.1} & \cellcolor{ImportantColor}3.64 &36.2 / 10 \\
\bottomrule
\end{tabular}
\label{tab:calvin}
%\vspace{-2mm}
\end{table}

\subsection{Generalization on CALVIN}\label{sec:calvin}
The CALVIN benchmark is built on the PyBullet~\cite{coumans2016pybullet} simulator and features a Franka Panda robot arm interacting with its scene. It includes 34 tasks across four environments (A, B, C, and D), each equipped with a desk, sliding door, drawer, LED button, lightbulb switch, and three colored blocks. The environments vary in desk texture and object placement. CALVIN provides 24 hours of unstructured play data, with 35\% annotated using natural language. 

We evaluate models under the zero-shot generalization setup: training on environments A, B, and C, and testing on D. Since CALVIN lacks a motion planner, models must directly predict robot pose trajectories. For a fair comparison, we follow the same evaluation protocol as Diffuser. Each method predicts at most 60 actions per sequence and is evaluated across three random seeds. We report the success rate and the average number of sequential tasks completed. Additionally, we compare DP3 and Diffuser in task completion time (in seconds) and the number of diffusion steps (NDS) required, with results shown in the final column. Other methods are excluded from this comparison, as they do not use diffusion-based policies.

As shown in Tab.~\ref{tab:calvin}, our method (last row) consistently outperforms existing approaches on the zero-shot long-horizon prediction task, achieving a roughly 10-point improvement in the success rate of the 5th key pose and producing trajectories with the highest average length. Furthermore, pairwise comparisons between DP3 and DP3+ADP (which integrates our proposed ADP into DP3), and between Diffuser and our method (Diffuser with ADP) highlight the effectiveness of ADP as a plug-and-play module. For DP3+ADP, we only adjust the input format and evaluation protocol to align with the CALVIN benchmark, without altering the core model. Overall, ADP substantially improves the generalization of DPs while also boosting generation efficiency, reducing computational time by approximately 25\%.

\begin{table*}[t]
\vspace{-2mm}
\footnotesize
%\caption{Evaluation of generalizability of our ADPro on RealWP with or without fine-tuning.}
\caption{Evaluation of generalizability of ADPro on RealWP for five continuous key pose prediction, with and without fine-tuning.}
\vspace{-2mm}
\renewcommand{\tabcolsep}{1.3mm}
\centering
\begin{tabular}{@{}l|cccccc|ccccccc@{}}
\toprule
\multirow{2}{*}{Methods}  & \multicolumn{5}{c}{with fine-tuning}& \multirow{2}{*}{Avg. Len}&\multicolumn{5}{c}{without fine-tuning}&\multirow{2}{*}{Avg. Len} & \multirow{2}{*}{Avg. Time(s)} \\
    & 1& 2  & 3 & 4  & 5 & &  1 & 2 &   3  & 4  &  5 & &  \\
\hline
Diffuser       & 66.0& 49.5  & 36.6 & 26.8  & 18.5 &1.97 &  38.8 & 26.5 & 16.3& 8.2  &  2.1  &0.92 &0.95 \\ 
\textbf{Ours}&  86.5 & 67.5  & 52.1 & 39.0 & 27.7 &\textbf{2.73}  &53.1 & 40.8 & 32.7& 24.5  &  16.3   &\textbf{1.18} &0.85  \\ 
\bottomrule
\end{tabular}
\label{tab:RealWorldPlay}
\vspace{-1mm}
\end{table*}

\begin{table*}[tbh!]
    \centering
    \renewcommand{\tabcolsep}{1.2mm}
    \caption{Comparison of grasp success and inference time across 4 Acronym object categories.}
    \vspace{-2mm}
    \resizebox{0.9\textwidth}{!}{%
    \begin{tabular}{@{}l|cc|cc|cc|cc@{}}
    \toprule
    & \multicolumn{2}{c|}{Bottle} & \multicolumn{2}{c|}{Laptop} & \multicolumn{2}{c|}{Book} & \multicolumn{2}{c}{ToyFigure} \\
    & Avg. Success & Avg. Time & Avg. Success & Avg. Time & Avg. Success & Avg. Time & Avg. Success & Avg. Time\\
    \midrule
    SEDF(S=20)   & 48.0 & 0.019 & 37.1 & 0.028 & 61.2 & 0.020  & 38.6 &  0.024 \\
    SEDF(S=120)  & 51.4 & 0.133 & 41.9 & 0.116 & 66.8 & 0.113  & \textbf{44.0} & 0.110  \\
    \textbf{SEDF+ADP(S=20)}  & \textbf{52.7} & 0.025 & \textbf{43.0} & 0.032 & \textbf{73.1} & 0.027  & \textbf{44.0} & 0.029\\
    \bottomrule
\end{tabular}
}
\label{tab:SEDF} 
\vspace{-3mm}
\end{table*}

\subsection{Generalization on Real-world Data}\label{sec:generalizability}
Real\_World\_Play~(RealWP)~\cite{mees23hulc2}  contains 9 hours of unstructured real-world data collected by experts using Franka Panda robots. The dataset covers over 25 distinct manipulation skills, which are highly diverse and challenging due to their unstructured and sometimes suboptimal nature. A total of 3,605 episodes are annotated with language descriptions of the robot’s behavior. Each episode includes RGB-D images from a static and a gripper-mounted camera, proprioceptive data, and a 7-DoF action. Since RealWP lacks a motion planner, we predict gripper pose trajectories, similar to CALVIN. To do this, we extract frames with significant motion changes as key poses using the algorithm from 3D Diffuser Actor. This process yields a trajectory of 6, 7, or 8 steps for each episode.

%ManiSkill~\cite{gu2023maniskill2} and 
To evaluate the generalization ability of ADP, we directly applied the model trained on CALVIN to predict five key poses sequentially. The accuracy of the five consecutive steps trajectory, shown on the right side of Tab.~\ref{tab:RealWorldPlay}, demonstrates that our method enhances generalization across tasks and environments, even without fine-tuning. We further fine-tuned the model using 80\% of the 3,605 episodes and evaluated it on the remaining 20\%. The results, presented on the left Tab.~\ref{tab:RealWorldPlay}, show that after fine-tuning, our ADPro substantially boosts the policy's one-step success rate by 20 percentage points.

\subsection{Ablations} \label{sec:ablation}

The proposed adaptive policy \textbf{can be seamlessly integrated into existing diffusion policies in a plug-and-play manner}. To assess its effectiveness in improving success rate and efficiency of manipulation, \textbf{we incorporated it into three baseline models: DP3, SE(3)-DiffusionFields}~(SEDF)~\cite{urain2023se}, and \textbf{Diffuser}. We then conducted a quantitative comparison of their results, as shown in Tab.~\ref{tab:calvin} and \ref{tab:SEDF}. 

For SEDF, we evaluate the pre-trained model on the Acronym dataset~\cite{eppner2021acronym}, which provides successful 6-DoF grasps for diverse ShapeNet \cite{chang2015shapenet} objects. 
Grasp success rates are evaluated in simulation using Isaac Gym~\cite{makoviychuk2isaac}, following the same evaluation protocol of~\cite{urain2023se}. We evaluate the model on four representative object categories: Bottle, Laptop, Book, and ToyFigure, each with the first 50 instances selected from the dataset. For each category, we report the average grasp success rate and the average generation time per grasp (in seconds), under both low-step (S=20) and full-step (S=120) diffusion settings. The results are summarized in Tab.~\ref{tab:SEDF}. Our method (SEDF+ADP) achieves either comparable or superior success rates compared to both the baseline SEDF (S=20) and the high-step SEDF (S=120), while achieving up to 5× faster grasp generation compared to SEDF (S=120). In addition, we evaluate the multi-object model on a diverse set of 61 objects, each from a distinct category, by executing 200 generated grasps per object.  
As shown in Fig.~\ref{fig:combined_dsteps}, integrating ADP consistently improves performance across all diffusion step settings. On average, our method achieves a 2.7\% increase in success rate compared to SEDF with the same steps.

Furthermore, to evaluate the effectiveness of the major component of our ADPro, we consider the following ablative versions of our model:  
\textbf{i) without~(w/o) initial noise constraint}~(INC), 
\textbf{ii) w/o spherical Gaussian constraint}~(SGC), 
\textbf{iii) w/o observation guidance}~(OG). 
Ablation studies on the RLBench and CALVIN benchmarks show that all three modules contribute effectively to model performance. Notably, removing the OG module leads to a substantial drop in performance and increased time costs, as shown in Tab.~\ref{tab:ablation}.

\begin{table}[tbh!]
%\footnotesize
\renewcommand{\tabcolsep}{1.3mm}
\vspace{-3mm}
\caption{\textbf{Ablation study.}  
    Our model significantly outperforms its counterparts that do not use the initial noise constraint, the Gaussian manifold constraint, or the observation guidance.
    }
\vspace{-2mm}
\centering
\begin{tabular}{@{}l|c|c|c|c@{}}
\toprule
    \multirow{2}{*}{Versions}& \multicolumn{2}{c|}{RLBench} & \multicolumn{2}{c}{CALVIN} \\
    &\cellcolor{ImportantColor}Avg. Succ. & Avg. Time(s) & \cellcolor{ImportantColor}Avg. Len & Avg. Time(s) \\
    \midrule
    w/o INC & \cellcolor{ImportantColor}82.4& 421.6 &\cellcolor{ImportantColor}3.60 &35.2\\
    w/o SGC & \cellcolor{ImportantColor}82.1& 418.4&\cellcolor{ImportantColor}3.57 &35.8 \\
    w/o OG & \cellcolor{ImportantColor}81.6& 488.7 &\cellcolor{ImportantColor}3.38 &49.3 \\
    ADPro & \cellcolor{ImportantColor}83.9& 435.8&\cellcolor{ImportantColor}3.64  &36.2 \\
    \bottomrule
    \end{tabular}
    \label{tab:ablation}
    \vspace{-2mm}
\end{table}

\iffalse
\subsection{Evaluation in the real world}
We validate our ADPro in learning manipulation tasks from real-world demonstrations across 12 tasks.  We use a Franka Emika robot equipped with an Azure Kinect RGB-D sensor at a front view.  Images are originally captured at $1280 \times 720$ resolution and downsampled to a resolution of $256 \times 256$.  During inference, we utilize the BiRRT~\cite{kuffner2000rrt} planner provided by the MoveIt! ROS package~\cite{coleman2014reducing} to reach the predicted poses.

\begin{wraptable}{r}{6.5cm}
    \begin{adjustbox}{width=\linewidth}
    \centering
    \tb{@{}c|c|c|c|c|c@{}}{0.6}{
     close & put & insert peg & insert peg & put & open \\
     box & duck & into hole & into torus & mouse & pen \\
    \midrule
    100 & 100 & 50 & 30 & 80 & 100 \\
    \midrule
     press & put & sort & stack & stack & put block \\
     stapler & grapes & rectangle & blocks & cups & in triangle \\
    \midrule
    90 & 90 & 50 & 20 & 40 & 90 \\
    \bottomrule
    }
    \end{adjustbox}
    \caption{\textbf{Multi-Task performance on real-world tasks.}}
    \label{tab:real}
\end{wraptable}
\fi

\section{Conclusion}\label{sec:Conclusion}
In this paper, we present ADPro, a novel and adaptive diffusion policy that leverages observation guidance from observed data and constrains the diffusion process to improve generalizability in robotic manipulation. Unlike traditional methods, ADPro can generate transferable actions for unseen tasks without retraining. By imposing a spherical manifold constraint, the diffusion trajectories exhibit reduced backtracking and improved efficiency. Additionally, by initializing actions using the FGR algorithm rather than random noise, ADPro benefits from more informed starting points. Extensive experiments demonstrate that our training-free approach outperforms existing methods in both generalization and efficiency, confirming the potential % and effectiveness 
of our approach.

The limitation of our proposed method is its reliance on both the end-effector and the scene point clouds as input, which may constrain its applicable scenarios. However, such data is readily available in some datasets and can also be feasibly acquired on real robotic platforms. In future work, we will explore more flexible forms of guidance to reduce dependency on specific input modalities.

% \fi
\IEEEpeerreviewmaketitle

\ifCLASSOPTIONcaptionsoff
  \newpage
\fi

\bibliographystyle{IEEEtran}
%{\small
%\bibliographystyle{plainnat}
%\bibliographystyle{unsrt}
\bibliography{ref}
%}

\begin{IEEEbiography}[{\includegraphics[width=0.90in,height=1.30in,clip,keepaspectratio]{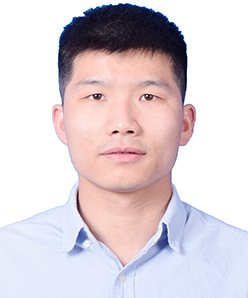}}]{Zezeng Li}
	received a B.S. degree from Beijing University of Technology (BJUT) in 2015 and a Ph.D. degree from Dalian University of Technology (DUT) in 2024. He is currently a postdoctoral fellow at the Ecole Centrale de Lyon (ECL).  His research interests include generative models and robotic manipulation.
\end{IEEEbiography}

\begin{IEEEbiography}[{\includegraphics[width=0.90in,height=1.30in,clip,keepaspectratio]{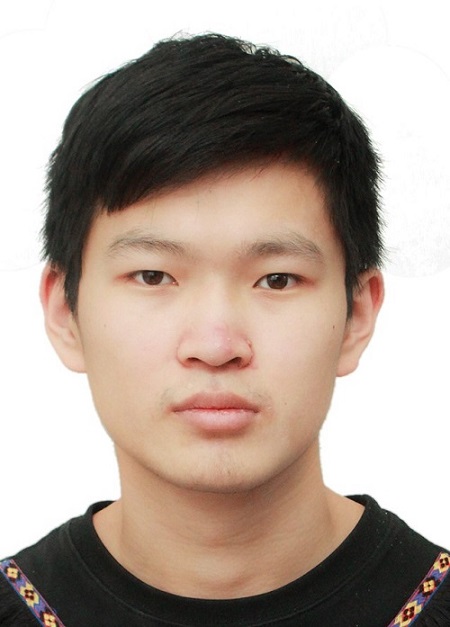}}]
    {Rui Yang} received a B.S. degree from Wuhan University in 2017 and an Engineering degree from École Centrale de Lyon (ECL) in 2020. He is currently pursuing a Ph.D. degree at LIRIS, École Centrale de Lyon. His research interests include continual learning and robotic manipulation.
\end{IEEEbiography}

\begin{IEEEbiography}[{\includegraphics[width=0.90in,height=1.30in,clip,keepaspectratio]{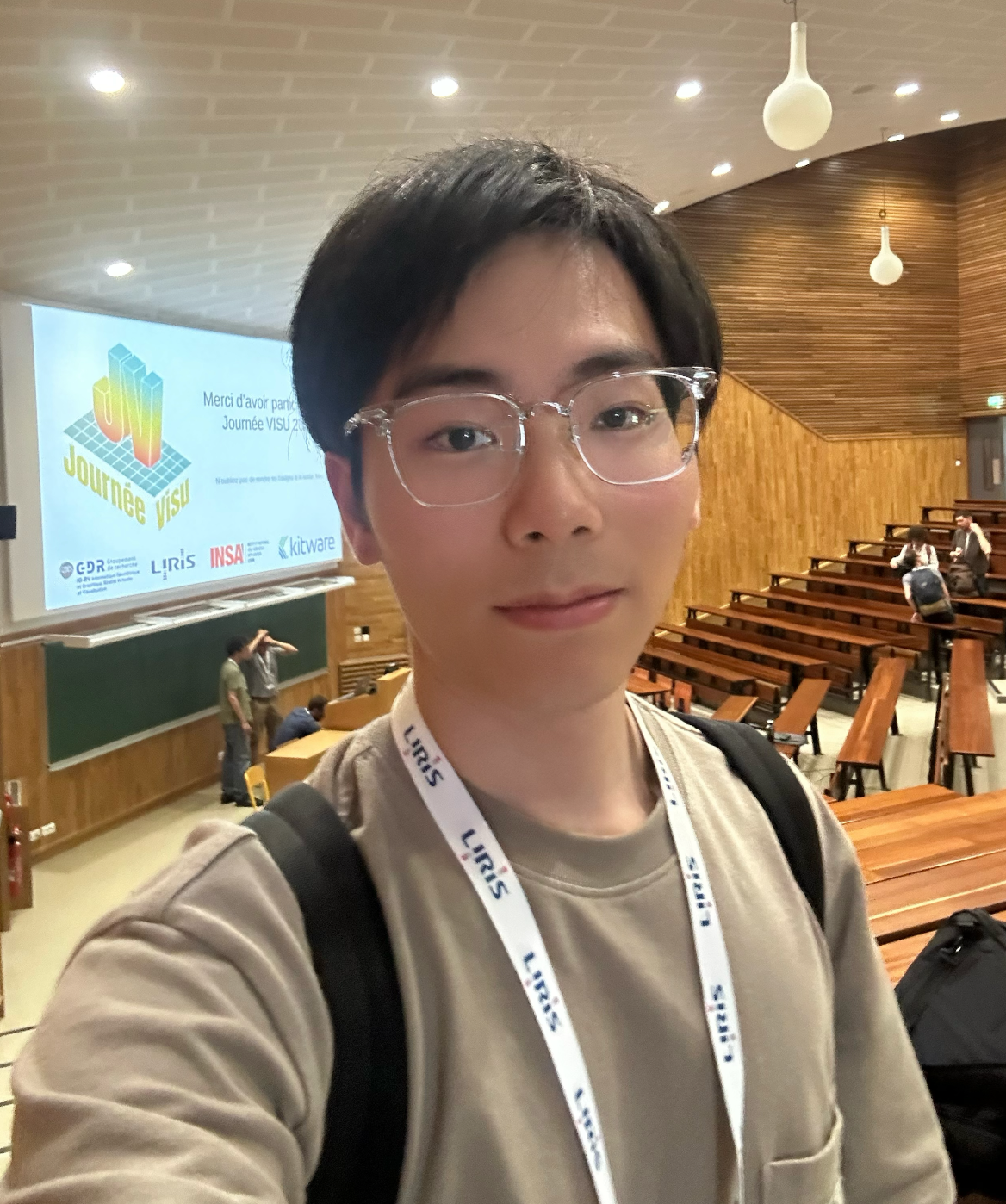}}]
    {Ruochen Chen} received an Engineering degree in Computer Science and Engineering (GI) and a Master’s degree in Machine Learning and Optimization of Complex Systems (AOS) from Université de Technologie de Compiègne (UTC) in 2022. He is currently pursuing a Ph.D. at LIRIS, École Centrale de Lyon, France. His research interests include deformable-object modeling and neural cloth and garment simulation.
\end{IEEEbiography}

\begin{IEEEbiography}[{\includegraphics[width=0.9in,height=1.3in,clip,keepaspectratio]{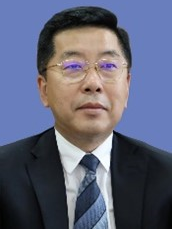}}]{Zhongxuan Luo}
received the B.S. degree from Jilin University in 1985 and the Ph.D. degree from Dalian University of Technology (DUT) in 1991. He has been a full professor with DUT since 1997, where he also serves as a president assistant. He is the director of the Liaoning Provincial Key Laboratory of Ubiquitous Network and Service Software. He is a member of the Software Engineering Professional Guidance Committee of the Ministry of Education of China. His research interests include computational geometry, computer vision and graphic imaging, and underwater agile robotics.\end{IEEEbiography}

\begin{IEEEbiography}[{\includegraphics[width=0.9in,height=1.60in,clip,keepaspectratio]{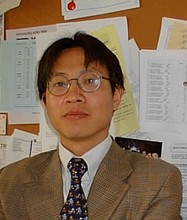}}]{Liming Chen}
	was awarded his B.Sc. degree in joint mathematics-computer science from the University of Nantes, France, in 1984, and his M.S. and Ph.D. degrees from the University of Paris 6, France, in 1986 and 1989. He first served as an Associate Professor with the Universite de Technologie de Compi`egne, before joining the Ecole Centrale de Lyon as a Professor in 1998, where he leads an Advanced Research Team in multimedia computing and pattern recognition. His current research interests include computer vision and multimedia, and in particular face analysis, image and video categorization, affective computing, and robotic manipulation. He is a Senior Member of the IEEE.
\end{IEEEbiography}

\end{document}